\pdfoutput=1
\documentclass[11pt]{article}

\usepackage[final]{acl}
\usepackage{algorithm}
\usepackage{amsmath}
\usepackage{amssymb} 
\usepackage{algpseudocode}
\usepackage{listings}
\usepackage{xcolor}
\definecolor{darkred}{rgb}{0.8, 0.2, 0.2}
\definecolor{darkgreen}{rgb}{0.2, 0.4, 0.2}
\usepackage{times}
\usepackage{latexsym}

\usepackage{marvosym}

\usepackage[T1]{fontenc}
\usepackage[utf8]{inputenc}

\usepackage{microtype}

\usepackage{inconsolata}

\usepackage{graphicx}

\title{KBAlign: Efficient Self Adaptation on Specific Textual Knowledge Bases}

\author{
 \textbf{Zheni Zeng\textsuperscript{1*}},
 \textbf{Yuxuan Chen\textsuperscript{1*}},
 \textbf{Shi Yu\textsuperscript{1}},
 \textbf{Ruobing Wang\textsuperscript{3}},
 \textbf{Yukun Yan\textsuperscript{1\Letter}},
 \\
 \textbf{Zhenghao Liu\textsuperscript{2}},
 \textbf{Shuo Wang\textsuperscript{1}},
 \textbf{Xu Han\textsuperscript{1}},
 \textbf{Zhiyuan Liu\textsuperscript{1}},
 \textbf{Maosong Sun\textsuperscript{1}},
\\
 \textsuperscript{1}Tsinghua University, Beijing, China
 \\
 \textsuperscript{2}Northeastern University, Shenyang, China
 \\
 \textsuperscript{3}University of Chinese Academy of Sciences, Beijing, China
 \\
   \href{mailto:email@domain}{yanyk.thu@gmail.com}
}

\begin{document}
\maketitle
\begin{abstract}
%Humans can utilize techniques to quickly acquire knowledge from specific materials in advance, such as creating self-assessment questions, enabling us to achieving related tasks more efficiently. In contrast, large language models (LLMs) usually relies on retrieval-augmented generation to exploit knowledge materials in an instant manner, or requires external signals such as human preference data and stronger LLM annotations to conduct knowledge adaptation. 
%To unleash the self-learning potential of LLMs, we propose KBAlign, an approach designed for efficient adaptation to textual knowledge bases. Our method utilizes iterative tuning with self-annotated data such as Q\&A pairs and revision suggestions, enabling the model to align with the specific knowledge domain efficiently, and perform better on downstream tasks involving knowledge content. 
%Experimental results on multiple datasets demonstrate the effectiveness of our approach, significantly boosting model performance in downstream tasks that require specific knowledge at a low cost. Notably, our approach achieves over 90\% of the performance improvement that can be obtained by using GPT-4-turbo annotation, while relying entirely on self-supervision. We release our experimental data, models, and process analyses to the community for further exploration
%~\footnote{~\url{https://github.com/thunlp/KBAlign}}.

Although retrieval-augmented generation (RAG) remains essential for knowledge-based question answering (KBQA), current paradigms face critical challenges under specific domains. Existing methods struggle with targeted adaptation on small-scale KBs: vanilla unsupervised training exhibits poor effectiveness, while fine-tuning incurs prohibitive costs of external signals. 
We present KBAlign, a self-supervised framework that enhances RAG systems through efficient model adaptation. Our key insight is to leverage the model's intrinsic capabilities for knowledge alignment through two innovative mechanisms: multi-grained self-annotation that captures global knowledge for data construction, and iterative tuning that accelerates convergence through self verification. This framework enables cost-effective model adaptation to specific textual KBs, without human supervision or external model assistance.
Experiments demonstrate that KBAlign can achieve 90\% of the performance gain obtained through GPT-4-supervised adaptation, while relying entirely on self-annotation of much smaller models. KBAlign significantly improves downstream QA accuracy across multiple domains with tiny costs, particularly benefiting scenarios requiring deep knowledge integration from specialized corpora. We release our experimental data, models, and process analyses to the community for further exploration~\footnote{~\url{https://github.com/thunlp/KBAlign}}.
\end{abstract}
\section{Introduction}

\begin{figure}[ht]
\centering
\includegraphics[width=1.1\linewidth]{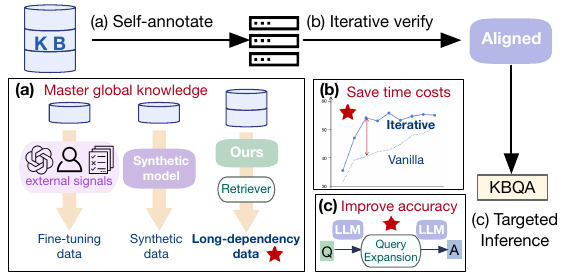}
\caption{KBAlign schematic. We design special self-annotation methods to help master global KB knowledge, conduct iterative verifying to save training time costs, and adopt targeted inference to improve accuracy.}
\label{fig:intro}
\end{figure}

Large language models (LLMs) have demonstrated their general capabilities across a wide range of downstream tasks~\citep{achiam2023gpt}, and the factual reliability of the models could be enhanced with common techniques such as retrieval augmented generation (RAG)~\citep{lewis2020retrieval}. When applied in specific domains, however, the adaptation of models to knowledge base (KB) materials remains a crucial strategy to further improve performance~\citep{ling2023domain}. Intuitively, adapted models align with the knowledge scope of the given KBs, rewriting the retrieval queries in a targetd manner and utilizing the KB information more fully. 
For example, when refering to the word ``LLM'', both general QA model and the retriever may regard it as an AI term, while the aligned model can supplement keywords to the initial query according to the scope of legal KBs, and then generates a disambiguated response with the help of a more precise retrieved context, in which ``LLM'' stands for master of laws.

%As shown in Fig.~\ref{fig:intro}, the aligned model supplements ``legal career'' keywords to the initial query according to the scope of KB, and then generates a disambiguated response with the help of a more precise retrieved context, in which ``LLM'' stands for master of laws.%例子再议

%while domain adaptation remains a crucial strategy to further enhance their performance~\citep{ling2023domain}. There are often quite specific needs for LLM adaptation in real-world scenarios, such as providing customized services based on user-specific document repositories, or plug-and-play integration of modules (e.g., operation manuals for new tools), in which the knowledge sources could be defined as textual knowledge bases. These scenarios often have smaller data amount and come with constraints such as confidentiality and convenience, leading to the inability to involve human annotation or rely on online large models, along with limited computational resources. In light of these constraints, retrieval augmented generation (RAG)~\citep{lewis2020retrieval} has emerged as a common approach to enhance the factual reliability of models. Considering that general retrieval methods do not always take advantage of specific knowledge, direct adaptation of models to specific KBs remains necessary~\citep{zhao2024chemdfm}. 

Existing adaptation methods usually construct domain models with large-scale training data~\citep{zhao2024chemdfm}, while there are quite specific needs in real-world scenarios corresponding to small-scale textual KBs, such as providing customized services based on user-specific document repositories, or plug-and-play integration of modules. In these cases, simple LM training on raw data may degrade performance, necessitating alternative approaches to align with specific domains~\citep{cheng2023adapting}. Targeted fine-tuning, on the other hand, typically involves the incorporation of external knowledge signals~\citep{tan2024large} to transform data into more structured tasks. When faced with constraints such as confidentiality, convenience, and limited computational resources, involving human annotation or relying on online large models becomes unpractictible. Therefore, a low-cost adaptation to small-scale KBs that does not rely on external supervisions is urgently needed.

%Direct training on small-scale raw data, however, may degrade performance, necessitating alternative approaches to align with specific domains. Common solutions try to transform data into more structured fine-tuning tasks such as reading comprehension~\citep{cheng2023adapting}, typically involve the incorporation of external knowledge signals (e.g., hand-crafted rules or GPT-4 annotations~\citep{tan2024large}). Nevertheless, it has been proven that LLMs can achieve self-improvement and refinement. For example, leveraging internal consistency and self-feedback~\citep{liang2024internal}, researchers have developed various filtering, correction, and preference training strategies to enhance the reliability of model outputs. Therefore, external knowledge signals are not necessary to some extent, and activating the potential of LLMs for autonomous adaptation is a better choice. Drawing an analogy to human learning process, RAG is similar to open-book tests in which students would conduct self-study in advance. To be specific, they quickly read the fundamental content of the books by themselves and can also verify their own answers. This may not be as good as spending time learning from external signals (e.g., related information from teachers), but is much more efficient. Therefore, rapid self-adaptation to specific knowledge materials is a useful capability.

Drawing an analogy to the human learning process, RAG is similar to open-book tests in which students could query KB materials. If they conduct self-study in advance, quickly grasping the fundamental content of the books themselves, the effectiveness and efficiency in tests can be improved. Based on this idea, we purpose KBAlign, a highly efficient self-adaptation approach tailored to specific KBs comprising self-improvement learning. 
Generally, we align the model with the small domain in a highly efficient and completely self-supervised manner. As shown in Fig.~\ref{fig:intro}, for \textbf{self annotation}, we organize the original KB materials in multiple grains and conduct self annotation to get instruction-response pairs that can cover various downstream task scenarios. For \textbf{iterative tuning}, we require the model to check its own responses and help modify common mistakes in the current stage for a faster convergence. Meanwhile, we conduct \textbf{targeted inference} in which strategies such as query expansion and confidence verification are adopted to refine the response.

Experiments on fact QA, long-form QA, and professional field test have shown the effectiveness of our method across different backbone models. With a low cost, KBAligned models master the general knowledge content from the KB and achieves significant performance improvements. Side experiments including ablation studies and performance curve analyses identify the most efficient self-annotated data amount and optimal training volumes, offering valuable guidance for effectively applying KBAlign in practical scenarios.

Our main contributions are as follows: (1) We propose KBAlign, a novel method for autonomous LLM adaptation tailored to textual KBs. It helps LLMs perform KB adaptation relying entirely on self annotations; (2) We provide empirical insights into efficient self-adaptation to KBs, offering practical parameters and settings for deploying KBAlign; (3) We conduct a comprehensive analysis of the proposed self-adaptation framework. Through a range of evaluation metrics and case studies, we identify the effectiveness of KBAlign and discuss the current limitations of our approach, highlighting areas for future improvement.

\section{Related Work}

\textbf{Domain Adaptation}.Though LLMs have shown their impressive capabilities in various scenarios~\citep{jablonka202314}, training methods for LLMs to adapt to certain domains still emerge in endlessly, due to the vertical application requirements. For domains with plenty of data resources, researchers directly take domain materials in pre-training~\citep{wang2023pre, madani2023large}. In more cases, they continue to train based on general LLMs or mix domain data with the general corpus~\citep{wu2023bloomberggpt}. To adopt domain knowledge in a more efficient way, format conversion and annotation are often performed~\citep{zhang2024chemllm} for fine-tuning. Some works focus on different settings for synthetic generation of QA data~\citep{soudani2024,ushio2023}, while with the development of annotation model capabilities, the impact of specific synthetic strategies diminishes significantly. More crucially, existing approaches predominantly focus on local information and ignore global knowledge in synthesis. 

%For efficiency reasons, domain LLMs usually use parameter-efficient tuning to reduce the required data amount and training time~\citep{ding2022delta}. However, the construction of domain data is sometimes still costly. In our case, we mainly focus on the data perspective and try to propose more efficient data construction methods.

\textbf{Knowledge Enhancement}. For some specific downstream requirements, there often exist high-quality knowledge materials (e.g., domain KBs, personal documents or records), of which the data amount is not enough for model tuning, and knowledge enhancement methods can help improve the performance. There are two mainstream solutions. The first one is to rely on the strong in-context learning capability of LLMs~\citep{dong2022survey}, and adopt RAG~\citep{lewis2020retrieval} to enhance the model. Apart from textual materials such as Internet passages, it is proven that integrating special KBs and tools is also a good approach to improve the model performance with specific knowledge~\citep{cui2023chatlaw, jin2024genegpt, qin2023tool}. To provide more useful information in context, strategies including designing better queries for retrieval are proposed~\citep{wang2023query2doc, qian2024memorag}. The second solution is to augment training data based on knowledge materials. LLMs can help synthesize data in more styles~\citep{sun2023sql} or convert the original data into formats more suitable for training~\citep{cheng2023adapting}. Our method is special, emphasizing the self annotation instead of introducing new LLMs into the system.

\textbf{Self Improvement}. There are some works exploring the self-improve capability of LLMs, most of which focus on the automatic generation and selection of reasoning steps for existing answers, being helpful in tuning~\citep{huang2023large} and inference~\citep{jiang2023self}. Self-play fine-tuning in an iterative manner~\citep{chen2024self} also unlocks the full potential of golden data. Even without the ground-truth answers, intern consistency of LLMs can be adopted as an important supervision signal that can achieve improvement~\citep{liang2024internal}. Nevertheless, challenge remains for human-like self improvement, such as how to self correct the reasoning process~\citep{huanglarge}. We observe the human learning process and design corresponding self-improvement methods.

\begin{figure*}[ht]
\centering
\includegraphics[width=1\linewidth]{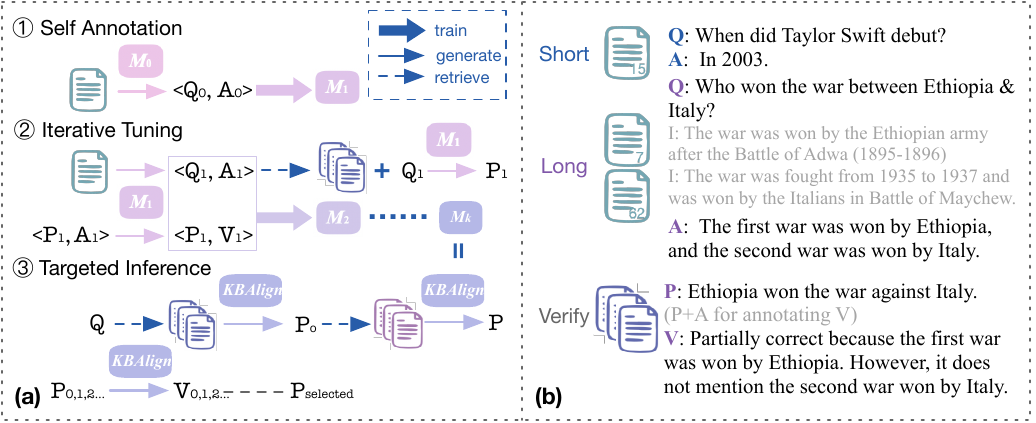}
\caption{(a) Details for the KBAlign framework; (b) Instances for different annotation strategies and tasks.}
\label{fig:task}
\end{figure*}

\section{Methodology}

\subsection{Task Setup}

%We define KB adaptation as the process where, given a knowledge base \( K \) (textual materials in our case), an original generative model \( M \), and a retriever \( R \), the goal is to efficiently enhance the grasp of the information contained within \( K \), thereby improving performance on a test set \( T \) based on \( K \). Common forms of \( T \) include knowledge question answering, multiple-choice tests, and other knowledge-intensive evaluations. The primary optimization objectives are to maximize performance scores on \( T \) while minimizing training time within constrained resources.

We define KB adaptation as the process in which, given a knowledge base \( K \) (textual materials in our case), an original generative model \( M \), and a retriever \( R \) for RAG, the goal is to efficiently align the models with the information in \( K \) without any external signals, thus improving the knowledge-intensive tasks based on \( K \). The optimization objectives are to maximize the performance scores in downstream tasks while minimizing training costs.

There are two common approaches utilizing \( K \) to enhance model performance: tuning-based and inference-based methods. Tuning-based methods involve generating tuning data of \( M\) from \( K \) using unsupervised techniques, or designing specific \( R \) for the current domain which is not covered in this work. Inference-based methods, on the other hand, focus on optimizing the retrieved content in the basic RAG setups, or post-processing the generated results to enhance relevance and accuracy. We now introduce our method which combines unsupervised tuning and RAG improvement, optimizing both the tuning and inference approaches. Examples are shown in Fig.~\ref{fig:task}.

\subsection{Self Annotation}

To learn the knowledge from KBs without any supervised data, we conduct self annotation with the backbone model \( M \) on the \( K \) text. To be specific, we choose a paragraph of golden context \( C_g \) as the knowledge source and require \( M \) to raise a set of questions \( Q \). We then supplement the related context \( C_R \) by the retriever \( R \), and ask \( M \) to annotate the answers \( A \) based on \( C_g + C_R \). When answering the questions with RAG, \( M \) sometimes fails due to the vague context provided by \(R\); while in the annotation process, \( A \) is comparably precise because of the ensured existence of \( C_g \) and our handcrafted keyword filters (e.g., questions should not mention pronouns such as ``in this paragraph''). 

%\( M \) is trained on <\( Q \), \( A \)> instead of directly on \( K \), which is similar with the self learning of human students converting and understanding the textbooks instead of directly reciting. 

Owing to the diverse forms and attributes of \( K \) and associated downstream tasks, we propose multi-grained annotation corresponding to different organization strategies for \( C \). The detailed process is shown in algorithm~\ref{alg:qa_train}.

\textbf{Short-dependency Annotation}. For downstream tasks that prioritize precise fact knowledge expressed in one specific paragraph, we employ this approach to simply divide \( K \) into fixed-length chunks, each with no more than $1,024$ words while keeping continuity of information across boundaries. One chunk is adopted directly as the annotation context \( C_g \).

\textbf{Long-Dependency Annotation}. Considering that real-world tasks often require a comprehensive understanding of multiple pieces of information at long distances in text, we design long-dependency annotation methods that split \( K \) into shorter segments with less than $256$ words. Several segments \( S_{1,...,n} \) with the same hierarchical directory, or with the highest embedding similarities across different directories, are concatenated as \( C_g \). When generating \( Q \), the model is required to raise questions that: (1) involve knowledge from different segments to emphasize the multi-hop reasoning capability; (2) are as vague as possible, corresponding to a series of information \( I_{1,...,n} \) annotated on \( S \), based on which a refined long-form answer \( A \) is generated to improve the integration capability.

\subsection{Iterative Tuning}

Apart from summarizing and self-questioning to help deep understanding, human students also take tests at each learning stage and strengthen the knowledge they have not yet mastered by correcting their answers. Similarly, we hope that the model can improve itself through self-verification in addition to understanding. The detailed process is also provided in algorithm~\ref{alg:qa_train}.

\textbf{Initial Tuning}. With the self-annotation <\( Q \), \( A \)> data, we tune \( M \) to get an initially adapted version. Due to the limitations of the retriever, the retrieved \( C_R \) in test scenario may differ from the annotation context \( C \) (in which the golden paragraph \( C_g \) must be included). Therefore, we randomly concatenate either \( C \) or \( C_R \)  with the question \( Q \) as the input. This mixed paradigm aims to bridge the gap between tuning and inference.%It is worth mentioning that a mixed paradigm has to be adopted to enhance the model generalization, that is, to provide a part of (50\% in our practice) retrieved context \( C_R \) which is different from the annotation \( C \) as the model input combined with \( Q \). This is to bridge the gap between tuning and inference in which RAG is not completely precise.

\textbf{Self-Verify Tuning}. %Based on the \( M_0 \) that is initially adapted on a part of the data <\( Q_0 \), \( A_0 \)>, we conduct RAG on \( Q_1 \) to get the predicted answer \( P_1 \) reflecting current capability of \( M_0 \). By providing the ground truth answer, the model verifies its own prediction and analyzes the wrong reason, which we name as \( V_1 \). In the next stage, we can then require the model to generate \( V_1 \) based on <\( Q_1 \), \( P_1 \)>. And so on, we generate the verifying data based on current performance, and conduct the Q\&A task and verify task at the same time in an iterative manner. In experiments, we use 25\% of verify and 75\% of Q\&A, and implement 2-3 iterations. 
We divide the annotated data into k parts <\(Q_{1,2,...k}\), \(A_{1,2,...k}\)>, and adapt the model with the first part <\( Q_1 \), \( A_1 \)> to get the initial version \( M_1 \). Using this model, we perform RAG inference in the second part \( Q_2 \) to obtain the predicted answer \( P_2 \) reflecting current capability of \( M_1 \). Given the ground-truth answer \( A_2 \), the model verifies its own prediction and analyzes the wrong reason, which we name \( V_2 \). In the next stage, we can then use <\( Q_2 \), \( P_2 \)> as input and \( V_2 \) as output to continue tuning \( M_0 \). And so on, we generate the verification data based on current performance, and conduct the Q\&A task and verify task at the same time in an iterative manner. In experiments, we use 25\% of verify and 75\% of Q\&A, and implement 2-3 iterations. 

\subsection{Targeted Inference}

We improve the downstream performance mainly by training the model to learn more specific knowledge. %Practically, there are also strategies that can help optimize the generation results in inference stage.
We also employ \textbf{Query Expansion} (QE) to refine the retrieval results in reference stage. To be specific, directly applying \( Q \) as the search query may miss useful information due to the short expression and the limitation of the retriever \( R \). Considering that our model has memorized the overall knowledge, it can provide a prediction \( P \) that is relevant to the KB content. We then expand the search query as \( Q \)+\( P \), and this may help make the retrieval results much better. 

The other strategy that can be used in reference is \textbf{Self Verification}, which is based on the capabilities learned in iterative verifying. For the generated \( P \), the model can check the correctness by itself. It should be emphasized that this is not the standard strategy setting in subsequent experiments, because it will increase the time cost, and it is also difficult for the model to correct the error after realizing it. However, the model can at least provide an uncertainty warning, or sample a new response when the confidence score is low when needed, which helps improve reliability.

\section{Experiment}
\subsection{Datasets and Models}

In order to evaluate the effectiveness of our method as comprehensively as possible, we use four datasets in the experiment, each could form a corresponding KB (from 0.41 to 21 M tokens). Details are displayed in section~\ref{sec:appendix}.

\textbf{LooGLE}~\citep{li2023loogle}. This is a long-text dataset, with textual materials that can be regarded as KBs and high-quality questions. We evaluate the specific knowledge memorizing capability of the model in this dataset.

\textbf{ASQA}~\citep{stelmakh2022asqa}. This is a long-form QA dataset. We evaluate the knowledge recall and organizing capability of the model in test set, and do not use any training data from it.

\textbf{JEC-QA}~\citep{zhong2020jec}. This is a legal multiple choice dataset in Chinese.  We evaluate the professional learning capability and instruction following in different inference formats. 

\textbf{BioASQ}~\footnote{\url{https://huggingface.co/datasets/kroshan/BioASQ}}. This is a biomedical question answering dataset. We evaluate the model's biomedical knowledge retrieval and reasoning capabilities.

We choose the following models as the backbone and comparison objects of the experiments:

\textbf{MiniCPM}~\citep{hu2024minicpm}. This refers to MiniCPM-2B which is one of the backbone models in our experiment. It is an end-side LLM gaining the instruction following ability during pre-training, and has achieved the best performance among lightweight LLMs on several datasets. Therefore, we believe that it has wide personal applications and is suitable for efficient adaptation scenarios.

\textbf{LLaMA-3.1}. This refers to LLaMA-3.1-8B-Instruct~\footnote{\url{https://huggingface.co/meta-llama/Llama-3.1-8B}} which is aligned from one of the most popular open-source model families. We choose it to evaluate whether our method is universally helpful when the backbone model becomes stronger.

\textbf{GPT series}. GPT-3.5 refers to GPT-3.5-turbo-0125~\footnote{\url{https://platform.openai.com/docs/models}} which is a representative closed-source LLM with stable performance and comprehensive capability. GPT-4o is an even stronger LLM. 

\textbf{LM}. This represents directly conducting the language modeling task to align the model with KBs. Knowledgeable text is segmented into 512-token-length paragraphs, and mixed with general instruction tuning data~\citep{ding2023enhancing} to keep the instruction following capability. 

\textbf{RAFT}~\citep{zhang2024raft}. This represents adapting language models to domain-specific RAG. Follow this method, we use GPT-4o to annotate data from the KBs, generating both Chain-of-Thought(CoT) reasoning and final answers to help the model focus on useful information while disregarding distractors. The annotated data used for supervised fine-tuning (SFT). We adopt this method as one of the baselines in our experiments. 

\subsection{Evaluation Metrics}

For LooGLE, ASQA and BioASQ, we consider the evaluation of the original dataset and decide to utilise the following metrics: (1) \textbf{Rule metrics}: F1 score, which measures the harmonic mean of precision and recall; Match score, which measures the recall of key elements in long-form answer; For JEC-QA, only precise prediction of options could be scored, regardless of whether the questions were single or multiple-choice. 

(2) \textbf{Similarity metrics}: BERT score~\citep{bert-score} calculates cosine similarity to assess semantic consistency, utilizing embeddings generated by the text2vec~\citep{text2vec} model from sentences; BLEU~\citep{papineni2002bleu}, ROUGE~\citep{lin2004rouge}, which are traditional text generation similarity metrics provided in ablation studies. 

(3) \textbf{Intelligent metrics}: semantic judgment by the representative OpenAI LLM, GPT-4o, is used to evaluate the quality of responses further. Detailed prompts are provided in Section~\ref{sec:appendix}.

\subsection{Other Settings}

We tune all parameters of MiniCPM, while conduct a parameter-efficient tuning for LLaMA-3.1, utilizing the LoRA~\citep{hu2021lora} strategy to reduce the need for computing power costs. In the test scenario, the chunks of KB materials are divided with less than 128 tokens, and the top 8 relevant chunks are provided.

\begin{table*}[ht]
\centering
\resizebox{\textwidth}{!}{
\begin{tabular}{l|ccc|ccc|ccc|ccc}
\hline
\textbf{Methods}        & \multicolumn{3}{c|}{\textbf{LooGLE}}                              & \multicolumn{3}{c|}{\textbf{ASQA}}        & \multicolumn{3}{c|}{\textbf{JEC-QA}}         &
\multicolumn{3}{c}{\textbf{BioASQ}}\\
                                   & F1     & BERT   & LLM   & Match  & BERT   & LLM   & Single & Multi & Total &
                                   F1   &
                                   BERT  &
                                   LLM\\ 
\hline

\multicolumn{13}{c}{\textbf{GPT series}}                                                                                          \\ 
\hline

GPT-3.5                          & 35.42  & 80.99 & 78.08  & 26.79   & 86.61 & 51.66  & 14.49   & 17.92  & 16.32  &  17.80  &  80.55 & 93.85\\
\ \ \ \textit{w/o} QE                            & 35.27  & 80.94 & 77.91  & 27.40   & 86.71 & 51.52  & 13.84   & 19.15  & 16.68  &  18.55  &  80.57 & 93.23\\

GPT-4o                          & 40.20  & 81.70 & 82.93  & \textbf{32.18}   & \textbf{87.14} & 67.76  & 21.95   & 26.42  & 24.33  &  31.73  &  81.31 & 94.15\\

\ \ \ \textit{w/o} QE                            & 40.21  & 81.71 & \textbf{83.29}  & 32.15   & 87.10 & \textbf{67.88}  & 20.11   & \textbf{27.36}  & 23.98  &  31.39  &  80.83 & 94.46\\
\hline
\multicolumn{13}{c}{\textbf{MiniCPM-2B}}                                                                                          \\ 
\hline
Vanilla RAG                            & 30.92  & 80.70 & 64.76  & 11.91   & 82.30 & 22.92  & 39.24   & 13,87  & 25.69  &  29.27  &  82.37 & 84.92\\

\ \ \ \textit{w/o} QE                            & 30.31  & 80.37 & 64.72  & 12.37   & 82.90 & 22.42  & 38.38   & 14.06  & 25.39  &  30.23  &  82.71 & 83.69\\

RAFT & 44.36 & 84.05 & 70.73 & 12.03 & 85.94 & 16.18 & 17.30 & 14.06 & 15.57 & 6.66 & 81.96 & 89.23\\
LM & 50.15  & 84.77 & 65.62 & 10.72  & 81.27 & 21.03 & 47.36 & 7.98 & 23.73 & 55.44 & 88.62 & 81.85\\

\textbf{Ours}             & 54.09  & 86.48 & 75.19   & 15.68    & 85.41  & 24.81  & \textbf{49.95}  & 9.94 & 28.91  &  61.38  &  89.95 & 87.69\\
\ \ \ \textit{\( \Delta \)} & \textcolor{darkred}{(+23.17)} & \textcolor{darkred}{(+5.78)} & \textcolor{darkred}{(+10.43)} & \textcolor{darkred}{(+3.77)}  & \textcolor{darkred}{(+3.11)} & \textcolor{darkred}{(+1.89)} & \textcolor{darkred}{(+10.71)} & \textcolor{darkgreen}{(-3.93)} & \textcolor{darkred}{(+3.22)} & \textcolor{darkred}{(+32.11)} & \textcolor{darkred}{(+7.58)} & \textcolor{darkred}{(+2.77)} \\
% ↑ ↓

\ \ \ \textit{w/o} QE            & 53.76 & 86.23 & 73.19  & 16.12  & 85.48  & 25.69  & 49.41 & 10.92 & \textbf{29.16} &  61.91  &  89.91& 89.54\\

\hline
\multicolumn{13}{c}{\textbf{LLaMA3.1-8B-Instruct}}                                                                                          \\ 
\hline
Vanilla RAG                                           & 40.46 & 81.57 & 77.15  & 20.21  & 84.93 & 37.28 & 22.70   & 24.66  & 23.73  & 27.96  & 81.55 & 92.62\\
\ \ \ \textit{w/o} QE                          & 39.94 & 81.50 & 77.08  & 20.03  & 85.14 & 35.64& 22.92   & 24.07  & 23.53  & 27.74  & 81.66 & 91.08\\

RAFT & 42.13 & 84.88 & 77.91 & 23.42 & 85.92 & 38.74 & 23.24 & 15.47 & 19.09 & 44.94 & 83.36 & 93.54\\
LM & 54.06 & 85.53 & 78.58 & 19.07 & 82.52 & 38.04 & 20.40 & 9.57 & 13.90 & 56.28 & 88.02 & 90.15\\

\textbf{Ours}               & \textbf{62.07} & \textbf{88.63} & 80.16  & 25.23  & 86.29 & 42.44  & 34.59  & 14.13  & 23.83 & 70.97  & 92.06 & 93.54\\
\ \ \ \textit{\( \Delta \)} & \textcolor{darkred}{(+21.61)}& \textcolor{darkred}{(+6.06)} & \textcolor{darkred}{(+2.85)} & \textcolor{darkred}{(+5.02)} & \textcolor{darkred}{(+1.36)} & \textcolor{darkred}{(+5.16)} & \textcolor{darkred}{(+11.89)} & \textcolor{darkgreen}{(-10.53)} & \textcolor{darkred}{(+0.10)} & \textcolor{darkred}{(+43.01)} & \textcolor{darkred}{(+10.51)} &
\textcolor{darkred}{(+0.92)}\\
\ \ \ \textit{w/o} QE              &  61.79 & 88.55 & 79.96  & 25.56  & 86.89 & 41.31 & 34.16  & 14.42  & 23.78 & \textbf{73.30}  & \textbf{92.72} & \textbf{94.48}\\ 
\hline

\hline
\end{tabular}
}
\caption{KBAlign adaptation experiments on LooGLE, ASQA, JEC-QA and BioASQ. We report average for 3 random seeds.}
\label{tab:main}
\end{table*}

Hyper-parameters, retrieval and speed-up settings are provided in Section~\ref{sec:appendix}.

\subsection{Result Analysis}

 \textbf{Time Costs}. We first estimate the time cost to prove the efficiency of our method. We provide the result on ASQA after scaling to the capacity of an A100 GPU: short-dependency annotation for 1k data items takes 30 min, long-dependency annotation for 1k data items takes 140 min, and iterative tuning process takes 160 min. Comparably, direct language modeling training takes 480 min, which is longer than the whole process of KBAlign. As for RAFT, it involves larger models and longer CoT responses requiring more annotation time, and the tuning time is controlled to be the same with us.
%TODO 此处补充一下language modeling方法的耗时

\textbf{Main Experiments}. Results for our experiments are shown in Table~\ref{tab:main}. We provide the GPT-series results, the initial version and the self-adapted version of both MiniCPM-2B and LLaMA-3.1-8B-Instruct on the four dataset. Overall, comparing the ``Ours'' lines with the corresponding vanilla RAG, we can see that KBAlign ensures an obvious improvement on most of the metrics, regardless of the dataset and the backbone model, showing its generalization and effectiveness. 

The simple language modeling helps align the model to several KBs and gets marginal promotion, while not always effective, further proving the necessity of self annotation. The advanced baseline RAFT relies on the quality of CoT reasoning, which requires quite large amount of training. When aligned with our high-efficient training setting, its effect is not always obvious. %Meanwhile, the instruction following capability of generating specific formats is also restricted, and therefore F1 and Match scores decrease.

\textbf{Task Differences}. Nevertheless, our strategies still produce differentiated effects in the four scenarios. For LooGLE which evaluates the master of precise local knowledge, self-annotated tuning brings a huge improvement (over 20\% on F1) and the adapted 2B model can surpass LLaMA-3.1-8B \& GPT-4o performance. For ASQA emphasizing long-form answer that covers global information, however, the improvement is comparably marginal (less than 5\% on Match). The first possible reason is that backbone models have already mastered the WikiPedia knowledge in pre-training, and extra adaptation is redundant. The second reason is the over emphasis of local knowledge in the responses, making QE strategy provide even more limited retrieval results. 

The same trend is also reflected in the single-choice and multiple-choice tests of JEC-QA. Our method easily surpasses some legal-domain models in the former (such as 40.8 single-choice score reported for 7B legal LLM~\citep{wan2024reformulating}), while in the latter the performance even declines slightly due to reasons such as the output format. This indicate the challenge of learning knowledge with a long information span and logical chain.

\begin{figure}[ht]
\centering
\includegraphics[width=1.\linewidth]{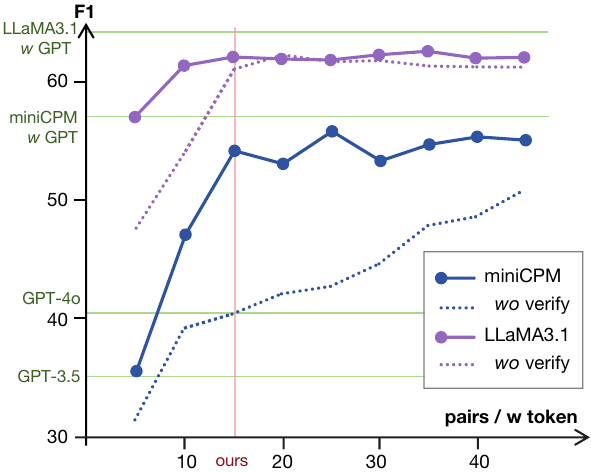}
\caption{The impact of training amount on LooGLE performance. `w GPT` refers to training with GPT-annotated data.}
\label{fig:numerical}
\end{figure}

\begin{figure}[ht]
\centering
\includegraphics[width=0.95\linewidth]{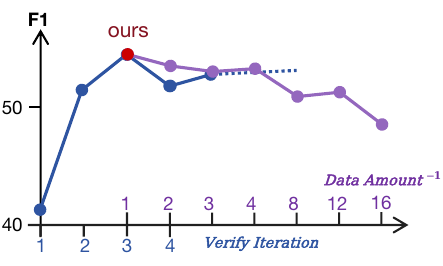}
\caption{The impact of iteration times and data amount for fixed training steps on LooGLE performance.}
\label{fig:iter}
\end{figure}

\textbf{Numerical Analysis}. We search the best values for key settings including the training steps, amount of data and iteration by evaluating checkpoints in process. From Fig.~\ref{fig:numerical} we can see, when learned on more QA pairs (only once), scores on LooGLE F1 (represents fact accuracy) for both backbone models improve. Interestingly, directly learning without iterative tuning (dotted curve) also displays a similar trend, while the tipping point for slowing growth comes much later. This reveals the possible mechanism of self-verify task, that is, to guide the model to focus more on the problems of current stage, so as to reach convergence faster. According to the curve, we choose to provide 15 data items per 10,000 tokens for LooGLE training, and increase the data density of ASQA due to the smaller KB scale. Besides, although the tuning phase usually reuses the same data for multiple epochs of training, we observe from Fig.~\ref{fig:iter} that using half of the data to tune 2 epoch brings a quite obvious score decrease. Consider that the inference time for data annotation is acceptable compared with the training time, we recommend annotating more data and tuning with only 1 epoch.

From Fig.~\ref{fig:iter} we can also observe the performance change for different number of iterations when conducting iterative tuning. With the self-verify data, the score first increases and then keeps comparably stable as the iterations increases, showing that the verification capability helps improve model performance on downstream QA, while requiring the data quality to be high enough with a certain granularity. From the curve we recommend conducting at least 3 iterations, while it depends on actual situation in practical implementation.

\begin{table}[ht]
\centering
\resizebox{\linewidth}{!}{
\begin{tabular}{lccccc}
\hline
\textbf{Methods}      & \textbf{F1}   & \textbf{BLEU} & \textbf{ROUGE} & \textbf{BERT}  & \textbf{LLM} \\ 
\hline
\multicolumn{6}{c}{\textbf{MiniCPM-2B}}                                                                                                                               \\

\hline
\textit{GPT Data}            & \textbf{56.92} & \textbf{20.72} & \textbf{52.30} & \textbf{87.16} & \underline{72.83} \\
\textbf{Ours}       & \underline{54.09} & 18.32 & \underline{49.75} & \underline{86.48} & \textbf{75.19} \\
\ \ \ \textit{w/o} know    & 52.16 & 14.71 & 47.85 & 85.75 & 69.35 \\
\ \ \ \textit{w/o} RAG      & 15.40 & 1.66 & 15.16 & 73.54 & 11.64 \\
\ \ \ \textit{w/o} QE      & 53.76 & \underline{18.55} & 49.61 & 86.23 & 73.19    \\
\ \ \ \textit{w/o} verify  & 42.69 & 15.95 & 39.57 & 82.74 & 72.37 \\
\hline
\multicolumn{6}{c}{\textbf{Llama3.1-8B-Instruct}}    \\
\hline
\textit{GPT Data}            & \textbf{64.97} & \textbf{26.41} & \textbf{59.85} & \textbf{89.56} & \textbf{80.21} \\
\textbf{Ours}       & \underline{62.07} & \underline{21.73} & \underline{57.34} & \underline{88.63} & \underline{80.16}\\
\ \ \ \textit{w/o} know    & 58.32 & 21.22 & 52.32 & 87.12 & 78.31 \\
\ \ \ \textit{w/o} RAG     & 14.75 & 0.70 & 14.22 & 74.1 & 15.89 \\
\ \ \ \textit{w/o} QE      & 61.79 & 21.60 & 57.09 & 88.55 & 79.96 \\

\ \ \ \textit{w/o} verify      & {61.76} & {20.94} & {57.20} & {88.53}  &  77.81\\
\hline

\end{tabular}}
\caption{Detailed results on LooGLE.}
\label{tab:loogle}
\end{table}

\begin{table}[ht]
\centering
\resizebox{\linewidth}{!}{
\begin{tabular}{lccccc}
\hline
\textbf{Methods}      & \textbf{Match} & \textbf{BLEU}  & \textbf{ROUGE} & \textbf{BERT}  & \textbf{LLM} \\ 
\hline
\multicolumn{6}{c}{\textbf{MiniCPM-2B}} \\
\hline
\textit{Golden}         & \textbf{18.90}                    & \textbf{4.39} & \textbf{26.89}         & \textbf{88.16}         & 23.43           \\
\textbf{Ours}        &\underline{15.68}  & \underline{2.67} & \underline{24.59}  & \underline{85.41}  & \textbf{24.81}           \\
\ \ \ \textit{w/o} long     & 13.34                & 1.18  & 21.26          & 81.91         & \underline{24.32}             \\
\ \ \ \textit{w/o} verify   & 14.28  & 2.22 & 24.32  & 84.02 & 23.20             \\
\hline
\multicolumn{6}{c}{\textbf{Llama3.1-8B-Instruct}}                                                                                                                    \\
\hline
\textit{Golden}  & \textbf{28.41} & \textbf{3.95} & \textbf{26.75}& \textbf{88.30} & \textbf{44.08}           \\
\textbf{Ours}  & \underline{25.23}  & \underline{3.43} & \underline{23.59} & \underline{86.29} & 42.44          \\
\ \ \ \textit{w/o} long     & 20.45            & 0.64  & 17.29          & 79.85         & \underline{43.95}             \\
\ \ \ \textit{w/o} verify   & 24.88            & 2.18   &22.01          & 83.32         & 35.14 \\

\hline
\end{tabular}}
\caption{Detailed results on ASQA.}
\label{tab:asqa}
\end{table}

\begin{figure*}[ht]
\centering
\includegraphics[width=1\linewidth]{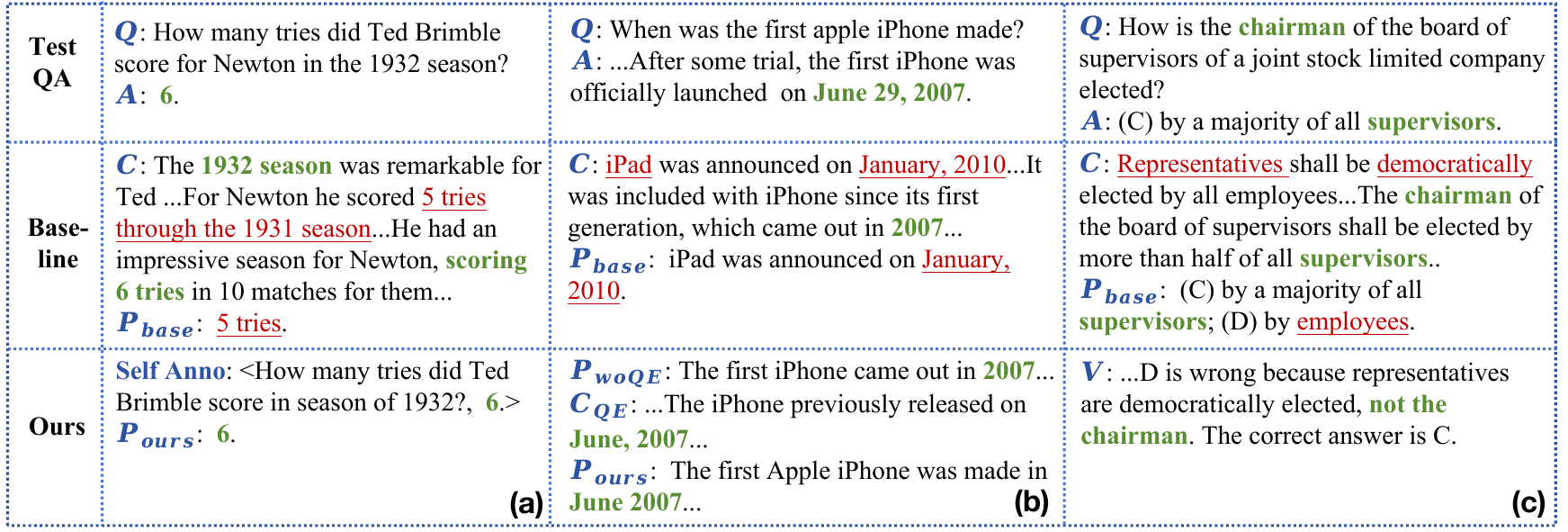}
\caption{Cases for KBAlign and baseline comparison. We display the translation for the Chinese JEC-QA task. The \textbf{bold} text and \underline{underlined} text providing correct and wrong information for the QA process.}
\label{fig:case}
\end{figure*}

\textbf{Ablation Study}. We assess the effectiveness of our strategies by side experiments in LooGLE and ASQA, and provide the results in Table~\ref{tab:loogle} and~\ref{tab:asqa}. To validate the quality of self annotated data, we try to replace the annotation model with  GPT-4-turbo (``GPT Data'') in LooGLE, and further replace the data with golden training set (``Golden'') in ASQA. We find that the eventual scores are not much higher than current setting, especially when compared with the vanilla setting without adaptation, proving the usefulness of self annotation.

By conducting ablation study on ASQA, we prove that the long-dependency annotation (``Ours'' vs. ``\textit{w/o} long'') plays a vital role, in which the comprehensive responses are expected. Considering the higher time cost (about 4 times of short-dependency), we discard this strategy to the local QA task in LooGLE. Meanwhile, self-verify tuning (``Ours'' vs. ``\textit{w/o} verify'') also helps improve the performance for both dataset by correcting errors of the current stage in a targeted manner.

To explore the mechanism of improvement more clearly, we also conduct a cross-validation on LooGLE (``\textit{w/o} know''), in which the amount of training data keeps the same while the exact information corresponding to the test questions are removed during self-annotation. We can see that the self adaptation still helps refine the performance, but worse than the complete setting. This indicates that domain knowledge from KB and task format is the main reason of the score rising, while the precise information related to the test data also helps.

\subsection{Case Studies}

We display typical cases in Fig.~\ref{fig:case} to explain the specific usefulness of our strategies. ``Baseline'' refers to MiniCPM-2B and ``ours'' refers to the adapted version of it. Overall, KBAlign achieves a general grasp of current KB, a better knowledge answering, and a reasonable confidence verification.

Cases (a) proves the effectiveness of learning knowledge from the self-annotated data. The base model fails to extract useful knowledge (scores in 1932 season) from the indirect context, while the model learns the precise knowledge during self adaptation. Case (b) shows that our model generates a decent prediction (in 2007) though the retriever fails to locate precise information from KB, and this prediction can then help find out useful knowledge with QE, therefore the model eventually provides an even better response (in June 2007). Further, from case (c) we can see that due to the self-verify task mixed into adaptation tuning, the model can check its own prediction and provide a hint of error or incompleteness. Though the verify reason is not always accurate or helpful for modification, it is still meaningful to provide a warning when the confidence is low. Meanwhile, we can also use the verify function as a self-selector for multiple sampling results. 

We also see some limitations when observing more cases. To be specific, the self-annotated contains some bias or error, and this may damage the model performance on related questions. Due to the concise language style of annotated data, our model tends to provide short responses in which some useful information may be discarded. QE strategy, in addition, does not always necessary. These negative instances remind us that we should continue to design better annotation and tuning strategies. More cases on different dataset and with various performance are provided in section~\ref{sec:appendix}.

\section{Discussion and Limitations}

In this paper, we introduced KBAlign, a highly efficient self-adaptive method tailored for specific KBs. During the tuning stage, inspired by human learning strategies such as summarizatio and self-reflection, we propose a combined long- and short-dependency annotation method, as well as an iterative tuning approach. These techniques enable low-cost targeted training data augmentation and efficient adaptation without requiring external supervision. In the inference stage, we enhance the model’s performance on KBQA tasks using query expansion and sampling-based self-verify strategies. Our approach demonstrates significant improvements across various datasets spanning different domains and formats. Additionally, detailed analysis provides empirical guidance regarding the best data amount required.

Still, our approach has some limitations: (1) \textbf{Global information}: While the current method excels in KBQA tasks, especially those focused on local information within the KB, it offers less support for tasks requiring comprehensive global information analysis. This suggests a need for more refined data annotation strategies. 

(2) \textbf{General Capability}: Training on small-scale targeted data can lead to a reduction in the model’s general domain abilities, such as instruction-following. Mixing specific KB data with general domain data, in fact, has proved to be helpful in our side experiment, which is displayed in section~\ref{sec:appendix}. However, this conflicts with our goal of minimizing adaptation time and cost. We may need to explore techniques like model plugins and routing selection to strike a better balance. 

(3) \textbf{Retriever Adaptation}: Given the strong influence of retrieval quality on QA performance found in our practice, it may be necessary to consider adapting the retriever during specific KB adaptations. Applying self-supervised strategies to retriever training could be a promising direction.

In future work, we aim to focus on adaptive performance enhancement in more complex scenarios, such as utilizing new tools. Additionally, we will explore the integration and collaboration of multiple models adapted to different subdomains.

\bibliography{custom}

\begin{thebibliography}{42}
\providecommand{\natexlab}[1]{#1}

\bibitem[{Achiam et~al.(2023)Achiam, Adler, Agarwal, Ahmad, Akkaya, Aleman, Almeida, Altenschmidt, Altman, Anadkat et~al.}]{achiam2023gpt}
Josh Achiam, Steven Adler, Sandhini Agarwal, Lama Ahmad, Ilge Akkaya, Florencia~Leoni Aleman, Diogo Almeida, Janko Altenschmidt, Sam Altman, Shyamal Anadkat, et~al. 2023.
\newblock Gpt-4 technical report.
\newblock \emph{arXiv preprint arXiv:2303.08774}.

\bibitem[{Chen et~al.(2024)Chen, Deng, Yuan, Ji, and Gu}]{chen2024self}
Zixiang Chen, Yihe Deng, Huizhuo Yuan, Kaixuan Ji, and Quanquan Gu. 2024.
\newblock Self-play fine-tuning converts weak language models to strong language models.
\newblock \emph{arXiv preprint arXiv:2401.01335}.

\bibitem[{Cheng et~al.(2023)Cheng, Huang, and Wei}]{cheng2023adapting}
Daixuan Cheng, Shaohan Huang, and Furu Wei. 2023.
\newblock Adapting large language models via reading comprehension.
\newblock \emph{arXiv preprint arXiv:2309.09530}.

\bibitem[{Cui et~al.(2023)Cui, Li, Yan, Chen, and Yuan}]{cui2023chatlaw}
Jiaxi Cui, Zongjian Li, Yang Yan, Bohua Chen, and Li~Yuan. 2023.
\newblock Chatlaw: Open-source legal large language model with integrated external knowledge bases.
\newblock \emph{arXiv preprint arXiv:2306.16092}.

\bibitem[{Ding et~al.(2023)Ding, Chen, Xu, Qin, Zheng, Hu, Liu, Sun, and Zhou}]{ding2023enhancing}
Ning Ding, Yulin Chen, Bokai Xu, Yujia Qin, Zhi Zheng, Shengding Hu, Zhiyuan Liu, Maosong Sun, and Bowen Zhou. 2023.
\newblock Enhancing chat language models by scaling high-quality instructional conversations.
\newblock \emph{arXiv preprint arXiv:2305.14233}.

\bibitem[{Dong et~al.(2022)Dong, Li, Dai, Zheng, Wu, Chang, Sun, Xu, and Sui}]{dong2022survey}
Qingxiu Dong, Lei Li, Damai Dai, Ce~Zheng, Zhiyong Wu, Baobao Chang, Xu~Sun, Jingjing Xu, and Zhifang Sui. 2022.
\newblock A survey on in-context learning.
\newblock \emph{arXiv preprint arXiv:2301.00234}.

\bibitem[{Gao et~al.(2023)Gao, Yen, Yu, and Chen}]{gao2023enabling}
Tianyu Gao, Howard Yen, Jiatong Yu, and Danqi Chen. 2023.
\newblock Enabling large language models to generate text with citations.
\newblock In \emph{Empirical Methods in Natural Language Processing (EMNLP)}.

\bibitem[{Hendrycks et~al.(2021)Hendrycks, Burns, Basart, Zou, Mazeika, Song, and Steinhardt}]{hendryckstest2021}
Dan Hendrycks, Collin Burns, Steven Basart, Andy Zou, Mantas Mazeika, Dawn Song, and Jacob Steinhardt. 2021.
\newblock Measuring massive multitask language understanding.
\newblock \emph{Proceedings of the International Conference on Learning Representations (ICLR)}.

\bibitem[{Heydar et~al.(2024)Heydar, Evangelos, and Faegheh}]{soudani2024}
Soudanim Heydar, Kanoulas Evangelos, and Hasibi Faegheh. 2024.
\newblock Fine tuning vs. retrieval augmented generation for less popular knowledge.
\newblock In \emph{Proceedings of the 2024 Annual International ACM SIGIR Conference on Research and Development in Information Retrieval in the Asia Pacific Region}.

\bibitem[{Hu et~al.(2021)Hu, Shen, Wallis, Allen-Zhu, Li, Wang, Wang, and Chen}]{hu2021lora}
Edward~J Hu, Yelong Shen, Phillip Wallis, Zeyuan Allen-Zhu, Yuanzhi Li, Shean Wang, Lu~Wang, and Weizhu Chen. 2021.
\newblock Lora: Low-rank adaptation of large language models.
\newblock \emph{arXiv preprint arXiv:2106.09685}.

\bibitem[{Hu et~al.(2024)Hu, Tu, Han, He, Cui, Long, Zheng, Fang, Huang, Zhao et~al.}]{hu2024minicpm}
Shengding Hu, Yuge Tu, Xu~Han, Chaoqun He, Ganqu Cui, Xiang Long, Zhi Zheng, Yewei Fang, Yuxiang Huang, Weilin Zhao, et~al. 2024.
\newblock Minicpm: Unveiling the potential of small language models with scalable training strategies.
\newblock \emph{arXiv preprint arXiv:2404.06395}.

\bibitem[{Huang et~al.(2023{\natexlab{a}})Huang, Gu, Hou, Wu, Wang, Yu, and Han}]{huang2023large}
Jiaxin Huang, Shixiang Gu, Le~Hou, Yuexin Wu, Xuezhi Wang, Hongkun Yu, and Jiawei Han. 2023{\natexlab{a}}.
\newblock Large language models can self-improve.
\newblock In \emph{Proceedings of the 2023 Conference on Empirical Methods in Natural Language Processing}, pages 1051--1068.

\bibitem[{Huang et~al.(2023{\natexlab{b}})Huang, Chen, Mishra, Zheng, Yu, Song, and Zhou}]{huanglarge}
Jie Huang, Xinyun Chen, Swaroop Mishra, Huaixiu~Steven Zheng, Adams~Wei Yu, Xinying Song, and Denny Zhou. 2023{\natexlab{b}}.
\newblock Large language models cannot self-correct reasoning yet.
\newblock In \emph{The Twelfth International Conference on Learning Representations}.

\bibitem[{Jablonka et~al.(2023)Jablonka, Ai, Al-Feghali, Badhwar, Bocarsly, Bran, Bringuier, Brinson, Choudhary, Circi et~al.}]{jablonka202314}
Kevin~Maik Jablonka, Qianxiang Ai, Alexander Al-Feghali, Shruti Badhwar, Joshua~D Bocarsly, Andres~M Bran, Stefan Bringuier, L~Catherine Brinson, Kamal Choudhary, Defne Circi, et~al. 2023.
\newblock 14 examples of how llms can transform materials science and chemistry: a reflection on a large language model hackathon.
\newblock \emph{Digital Discovery}, 2(5):1233--1250.

\bibitem[{Jiang et~al.(2023)Jiang, Dong, Wang, Zheng, Shang, Li, Jin, and Jiao}]{jiang2023self}
Xue Jiang, Yihong Dong, Lecheng Wang, Fang Zheng, Qiwei Shang, Ge~Li, Zhi Jin, and Wenpin Jiao. 2023.
\newblock Self-planning code generation with large language models.
\newblock \emph{ACM Transactions on Software Engineering and Methodology}.

\bibitem[{Jin et~al.(2024)Jin, Yang, Chen, and Lu}]{jin2024genegpt}
Qiao Jin, Yifan Yang, Qingyu Chen, and Zhiyong Lu. 2024.
\newblock Genegpt: Augmenting large language models with domain tools for improved access to biomedical information.
\newblock \emph{Bioinformatics}, 40(2):btae075.

\bibitem[{Kwon et~al.(2023)Kwon, Li, Zhuang, Sheng, Zheng, Yu, Gonzalez, Zhang, and Stoica}]{kwon2023efficient}
Woosuk Kwon, Zhuohan Li, Siyuan Zhuang, Ying Sheng, Lianmin Zheng, Cody~Hao Yu, Joseph~E. Gonzalez, Hao Zhang, and Ion Stoica. 2023.
\newblock Efficient memory management for large language model serving with pagedattention.
\newblock In \emph{Proceedings of the ACM SIGOPS 29th Symposium on Operating Systems Principles}.

\bibitem[{Lewis et~al.(2020)Lewis, Perez, Piktus, Petroni, Karpukhin, Goyal, K{\"u}ttler, Lewis, Yih, Rockt{\"a}schel et~al.}]{lewis2020retrieval}
Patrick Lewis, Ethan Perez, Aleksandra Piktus, Fabio Petroni, Vladimir Karpukhin, Naman Goyal, Heinrich K{\"u}ttler, Mike Lewis, Wen-tau Yih, Tim Rockt{\"a}schel, et~al. 2020.
\newblock Retrieval-augmented generation for knowledge-intensive nlp tasks.
\newblock \emph{Advances in Neural Information Processing Systems}, 33:9459--9474.

\bibitem[{Li et~al.(2023)Li, Wang, Zheng, and Zhang}]{li2023loogle}
Jiaqi Li, Mengmeng Wang, Zilong Zheng, and Muhan Zhang. 2023.
\newblock Loogle: Can long-context language models understand long contexts?
\newblock \emph{arXiv preprint arXiv:2311.04939}.

\bibitem[{Liang et~al.(2024)Liang, Song, Zheng, Wang, Yu, Li, Li, Xiong, and Li}]{liang2024internal}
Xun Liang, Shichao Song, Zifan Zheng, Hanyu Wang, Qingchen Yu, Xunkai Li, Rong-Hua Li, Feiyu Xiong, and Zhiyu Li. 2024.
\newblock Internal consistency and self-feedback in large language models: A survey.
\newblock \emph{arXiv preprint arXiv:2407.14507}.

\bibitem[{Lin(2004)}]{lin2004rouge}
Chin-Yew Lin. 2004.
\newblock Rouge: A package for automatic evaluation of summaries.
\newblock In \emph{Text summarization branches out}, pages 74--81.

\bibitem[{Ling et~al.(2023)Ling, Zhao, Lu, Deng, Zheng, Wang, Chowdhury, Li, Cui, Zhang et~al.}]{ling2023domain}
Chen Ling, Xujiang Zhao, Jiaying Lu, Chengyuan Deng, Can Zheng, Junxiang Wang, Tanmoy Chowdhury, Yun Li, Hejie Cui, Xuchao Zhang, et~al. 2023.
\newblock Domain specialization as the key to make large language models disruptive: A comprehensive survey.
\newblock \emph{arXiv preprint arXiv:2305.18703}.

\bibitem[{Madani et~al.(2023)Madani, Krause, Greene, Subramanian, Mohr, Holton, Olmos, Xiong, Sun, Socher et~al.}]{madani2023large}
Ali Madani, Ben Krause, Eric~R Greene, Subu Subramanian, Benjamin~P Mohr, James~M Holton, Jose~Luis Olmos, Caiming Xiong, Zachary~Z Sun, Richard Socher, et~al. 2023.
\newblock Large language models generate functional protein sequences across diverse families.
\newblock \emph{Nature Biotechnology}, 41(8):1099--1106.

\bibitem[{Papineni et~al.(2002)Papineni, Roukos, Ward, and Zhu}]{papineni2002bleu}
Kishore Papineni, Salim Roukos, Todd Ward, and Wei-Jing Zhu. 2002.
\newblock Bleu: a method for automatic evaluation of machine translation.
\newblock In \emph{Proceedings of the 40th annual meeting of the Association for Computational Linguistics}, pages 311--318.

\bibitem[{Qian et~al.(2024)Qian, Zhang, Liu, Mao, and Dou}]{qian2024memorag}
Hongjin Qian, Peitian Zhang, Zheng Liu, Kelong Mao, and Zhicheng Dou. 2024.
\newblock Memorag: Moving towards next-gen rag via memory-inspired knowledge discovery.
\newblock \emph{arXiv preprint arXiv:2409.05591}.

\bibitem[{Qin et~al.(2023)Qin, Hu, Lin, Chen, Ding, Cui, Zeng, and Huang}]{qin2023tool}
Yujia Qin, Shengding Hu, Yankai Lin, Weize~Chen Chen, Ning Ding, Ganqu Cui, Zheni Zeng, and Yufei Huang. 2023.
\newblock Tool learning with foundation models.
\newblock \emph{arXiv preprint arXiv:2304.08354}.

\bibitem[{Stelmakh et~al.(2022)Stelmakh, Luan, Dhingra, and Chang}]{stelmakh2022asqa}
Ivan Stelmakh, Yi~Luan, Bhuwan Dhingra, and Ming-Wei Chang. 2022.
\newblock Asqa: Factoid questions meet long-form answers.
\newblock \emph{arXiv preprint arXiv:2204.06092}.

\bibitem[{Sun et~al.(2023)Sun, Arik, Muzio, Miculicich, Gundabathula, Yin, Dai, Nakhost, Sinha, Wang et~al.}]{sun2023sql}
Ruoxi Sun, Sercan~{\"O} Arik, Alex Muzio, Lesly Miculicich, Satya Gundabathula, Pengcheng Yin, Hanjun Dai, Hootan Nakhost, Rajarishi Sinha, Zifeng Wang, et~al. 2023.
\newblock Sql-palm: Improved large language model adaptation for text-to-sql (extended).
\newblock \emph{arXiv preprint arXiv:2306.00739}.

\bibitem[{Tan et~al.(2024)Tan, Beigi, Wang, Guo, Bhattacharjee, Jiang, Karami, Li, Cheng, and Liu}]{tan2024large}
Zhen Tan, Alimohammad Beigi, Song Wang, Ruocheng Guo, Amrita Bhattacharjee, Bohan Jiang, Mansooreh Karami, Jundong Li, Lu~Cheng, and Huan Liu. 2024.
\newblock Large language models for data annotation: A survey.
\newblock \emph{arXiv preprint arXiv:2402.13446}.

\bibitem[{Ushio et~al.(2023)Ushio, Fernando, and Camacho-Collados}]{ushio2023}
Asahi Ushio, Alva-Manchego Fernando, and Jose. Camacho-Collados. 2023.
\newblock An empirical comparison of lm-based question and answer generation methods.
\newblock In \emph{Findings of the Association for Computational Linguistics: ACL}.

\bibitem[{Wan et~al.(2024)Wan, Zhang, Wang, Cheng, and Kurohashi}]{wan2024reformulating}
Zhen Wan, Yating Zhang, Yexiang Wang, Fei Cheng, and Sadao Kurohashi. 2024.
\newblock Reformulating domain adaptation of large language models as adapt-retrieve-revise: A case study on chinese legal domain.
\newblock In \emph{Findings of the Association for Computational Linguistics ACL 2024}, pages 5030--5041.

\bibitem[{Wang et~al.(2023{\natexlab{a}})Wang, Xie, Pei, Chen, Tiwari, Li, and Fu}]{wang2023pre}
Benyou Wang, Qianqian Xie, Jiahuan Pei, Zhihong Chen, Prayag Tiwari, Zhao Li, and Jie Fu. 2023{\natexlab{a}}.
\newblock Pre-trained language models in biomedical domain: A systematic survey.
\newblock \emph{ACM Computing Surveys}, 56(3):1--52.

\bibitem[{Wang et~al.(2023{\natexlab{b}})Wang, Yang, and Wei}]{wang2023query2doc}
Liang Wang, Nan Yang, and Furu Wei. 2023{\natexlab{b}}.
\newblock Query2doc: Query expansion with large language models.
\newblock In \emph{Proceedings of the 2023 Conference on Empirical Methods in Natural Language Processing}, pages 9414--9423.

\bibitem[{Wu et~al.(2023)Wu, Irsoy, Lu, Dabravolski, Dredze, Gehrmann, Kambadur, Rosenberg, and Mann}]{wu2023bloomberggpt}
Shijie Wu, Ozan Irsoy, Steven Lu, Vadim Dabravolski, Mark Dredze, Sebastian Gehrmann, Prabhanjan Kambadur, David Rosenberg, and Gideon Mann. 2023.
\newblock Bloomberggpt: A large language model for finance.
\newblock \emph{arXiv preprint arXiv:2303.17564}.

\bibitem[{Xiao et~al.(2024)Xiao, Liu, Zhang, Muennighoff, Lian, and Nie}]{bge_embedding}
Shitao Xiao, Zheng Liu, Peitian Zhang, Niklas Muennighoff, Defu Lian, and Jian-Yun Nie. 2024.
\newblock \href {https://doi.org/10.1145/3626772.3657878} {C-pack: Packed resources for general chinese embeddings}.
\newblock In \emph{Proceedings of the 47th International ACM SIGIR Conference on Research and Development in Information Retrieval}, SIGIR '24, page 641–649, New York, NY, USA. Association for Computing Machinery.

\bibitem[{Xu(2023)}]{text2vec}
Ming Xu. 2023.
\newblock \href {https://github.com/shibing624/text2vec} {text2vec: A tool for text to vector}.
\newblock \emph{Software}.

\bibitem[{Zhang et~al.(2024{\natexlab{a}})Zhang, Liu, Tan, Chen, Yan, Yan, Li, Huang, Yue, Zhou et~al.}]{zhang2024chemllm}
Di~Zhang, Wei Liu, Qian Tan, Jingdan Chen, Hang Yan, Yuliang Yan, Jiatong Li, Weiran Huang, Xiangyu Yue, Dongzhan Zhou, et~al. 2024{\natexlab{a}}.
\newblock Chemllm: A chemical large language model.
\newblock \emph{arXiv preprint arXiv:2402.06852}.

\bibitem[{Zhang et~al.(2024{\natexlab{b}})Zhang, Patil, Jain, Shen, Zaharia, Stoica, and Gonzalez}]{zhang2024raft}
Tianjun Zhang, Shishir~G Patil, Naman Jain, Sheng Shen, Matei Zaharia, Ion Stoica, and Joseph~E Gonzalez. 2024{\natexlab{b}}.
\newblock Raft: Adapting language model to domain specific rag.
\newblock \emph{arXiv preprint arXiv:2403.10131}.

\bibitem[{Zhang* et~al.(2020)Zhang*, Kishore*, Wu*, Weinberger, and Artzi}]{bert-score}
Tianyi Zhang*, Varsha Kishore*, Felix Wu*, Kilian~Q. Weinberger, and Yoav Artzi. 2020.
\newblock Bertscore: Evaluating text generation with bert.
\newblock In \emph{International Conference on Learning Representations}.

\bibitem[{Zhao et~al.(2024)Zhao, Ma, Chen, Sun, Li, Xu, Zhu, Zhu, Fan, Shen et~al.}]{zhao2024chemdfm}
Zihan Zhao, Da~Ma, Lu~Chen, Liangtai Sun, Zihao Li, Hongshen Xu, Zichen Zhu, Su~Zhu, Shuai Fan, Guodong Shen, et~al. 2024.
\newblock Chemdfm: Dialogue foundation model for chemistry.
\newblock \emph{arXiv preprint arXiv:2401.14818}.

\bibitem[{Zheng et~al.(2024)Zheng, Zhang, Zhang, YeYanhan, and Luo}]{zheng2024llamafactory}
Yaowei Zheng, Richong Zhang, Junhao Zhang, YeYanhan YeYanhan, and Zheyan Luo. 2024.
\newblock Llamafactory: Unified efficient fine-tuning of 100+ language models.
\newblock In \emph{Proceedings of the 62nd Annual Meeting of the Association for Computational Linguistics (Volume 3: System Demonstrations)}, pages 400--410.

\bibitem[{Zhong et~al.(2020)Zhong, Xiao, Tu, Zhang, Liu, and Sun}]{zhong2020jec}
Haoxi Zhong, Chaojun Xiao, Cunchao Tu, Tianyang Zhang, Zhiyuan Liu, and Maosong Sun. 2020.
\newblock Jec-qa: a legal-domain question answering dataset.
\newblock In \emph{Proceedings of the AAAI conference on artificial intelligence}, volume~34, pages 9701--9708.

\end{thebibliography}

\appendix

\section{Appendix}
\label{sec:appendix}

\subsection{Dataset Details}

\textbf{LooGLE}~\citep{li2023loogle}. We use the short-dependency data in LooGLE for retrofitting, combining altogether 2.2M tokens of text as \( K \), and the corresponding 1,951 Q\&A pairs for test. 

\textbf{ASQA}~\citep{stelmakh2022asqa}. For each question, there exists several related segments of text from WikiPedia, and a comprehensive long answer that covers much information from them. We collect 794 Q\&A pairs for test, their targeted segments and other related passages from WikiPedia, and get 1.8M tokens of text as \( K \). 

\textbf{JEC-QA}~\citep{zhong2020jec}. Related laws and reference books are seen as \( K \), including 21M tokens of text. The train set has also not been used for adaptation, and the test set contains 1,985 of multiple choices. 

\textbf{BioASQ}~\footnote{\url{https://huggingface.co/datasets/kroshan/BioASQ}}. This contains 324 English questions for testing, each annotated with relevant documents, snippets, and both exact and ideal answers. We utilize 0.41M tokens of text as \( K \).

\subsection{Detailed Settings}

For hyper-parameter settings, we conduct a grid search in the vicinity based on the empirical values provided in the sample code of the MiniCPM-2B model, and finally determine batch size as $8$ and learning rate as $1e-5$. Other settings include warm-up steps as $50$ and weight decay as $0.1$. For the LLaMA-3.1-8B-Instruct model, we adopt a parameter-efficient tuning approach using the LoRA strategy, with alpha as $16$ and rank as $8$.  Other settings include a cosine learning rate scheduler with a warm-up ratio of $0.1$ and a weight decay of $0.1$.

For the training process, we adopt the mixed-precision training with the BMTrain~\footnote{\url{https://github.com/OpenBMB/ModelCenter?tab=readme-ov-file}} and LLaMA Factory~\citep{zheng2024llamafactory} framework to speed up. 

For the inference process in both annotation and test, we adopt the bge-large-en-v1.5 model for English materials and bge-base-zh-v1.5 for Chinese~\citep{bge_embedding} as the basic retriever of RAG. To ensure continuity of information, we apply an overlap rate of 15\% between consecutive chunks. We adopt vLLM~\citep{kwon2023efficient} to speed up inference.

\subsection{Prompts}

Below are the prompt templates used for self annotation. For short-dependency annotation, we directly generate by:

\begin{lstlisting}
You are a master of extracting questions and answers from text. 
Based on the provided content, construct five questions and answers 
that should be directly based on the text content, separated by line breaks. 
Please ensure that the expression of the question clearly points to 
the specific information in the text, and avoid using vague or overly 
broad references. At the same time, emphasize direct references or 
specific details in the text to increase the accuracy and depth of the problem. 
The questions should be answerable in a few words. 
Output question and answer alternately on each line.
Content: {content}
Response:
\end{lstlisting}

For long-dependency annotation, we first generate questions:

\begin{lstlisting}
You will receive a document. Please generate 3 generalizable, 
ambiguous questions based on the document content. The questions 
should align with the themes of the document. Separate the questions 
by line breaks.
document: {document}
output:
\end{lstlisting}

Based on the questions, we then annotate the related information:

\begin{lstlisting}
You will receive a document and a question. Please provide an answer 
to the question based on the document information. If unable to answer, 
return 'none'; otherwise, output the answer directly.
document: {document}
question: {question}
output:
\end{lstlisting}

Last, we refine the information to get the answer:
\begin{lstlisting}
You will receive a concatenated answer from multiple sources. 
Please refine and optimize the expression to make it smoother. 
Output the final answer directly without unnecessary explanation.
question: {question}
answer: {answer}
output:
\end{lstlisting}

In the iterative tuning phase, we self-verified by:
\begin{lstlisting}
You are a teacher evaluating student responses. Remember:
1. If the student's response fully aligns with the golden answer, start your response with 'The student's response is correct because'.
2. Otherwise, start your response with 'The student response is wrong because', and provide the ERROR TYPE!!! (e.g., does not answer the question directly, provides totally wrong information, provides only part of the information, provides unrelated information)
3. Notice! You are NOT ALLOWED to directly point out the correct answer in your verification. You are NOT ALLOWED to directly point out the correct answer in your verification. You are NOT ALLOWED to directly point out the correct answer in your verification. You should only tell me the correctness and the error type.
Now here are the materials:
Reference: {reference}
Question: {question}
Golden Answer: {golden_answer}
Student Response: {student_response}
Please generate your verification. You should start with the judgement, and then EXPLAIN the reason / the error type.
\end{lstlisting}

Below is the prompt template used for downstream QA tasks:

\begin{lstlisting}
You are an expert who has read a lot of knowledge base. 
Please answer the question according to the content of the KB. 
<KB_{kb_id}> You can refer to some segments from the KB to help 
you answer the question. 
References:{references}
Now the question is: {question}
{dataset_prompt}
\end{lstlisting}

For different datasets, we change the \texttt{dataset\_prompt} to adjust the output style. Specifically, we refer to ALCE~\citep{gao2023enabling} when designing the ASQA prompt.

\begin{lstlisting}
(*@\textbf{LooGLE}@*): Please answer this question.

(*@\textbf{ASQA}@*): Write an accurate, engaging, and concise answer for the given question. Use an unbiased and journalistic tone.

(*@\textbf{JEC-QA}@*): The answer may be multiple or single, so be sure to choose all the correct options.
\end{lstlisting}

Below is the prompt template used for LLM evaluation~\citep{li2023loogle}:

\begin{lstlisting}
Given one question, there is a groundtruth and a predict_answer. 
Please decide whether they are the same or not in semantic. 
Please only output 'True' or 'False'.
Question: {question}
groundtruth = {ground_truth}
predict_answer = {predict}
\end{lstlisting}

\subsection{Supplementary Case}

\begin{figure*}[ht]
\centering
\includegraphics[width=1\linewidth]{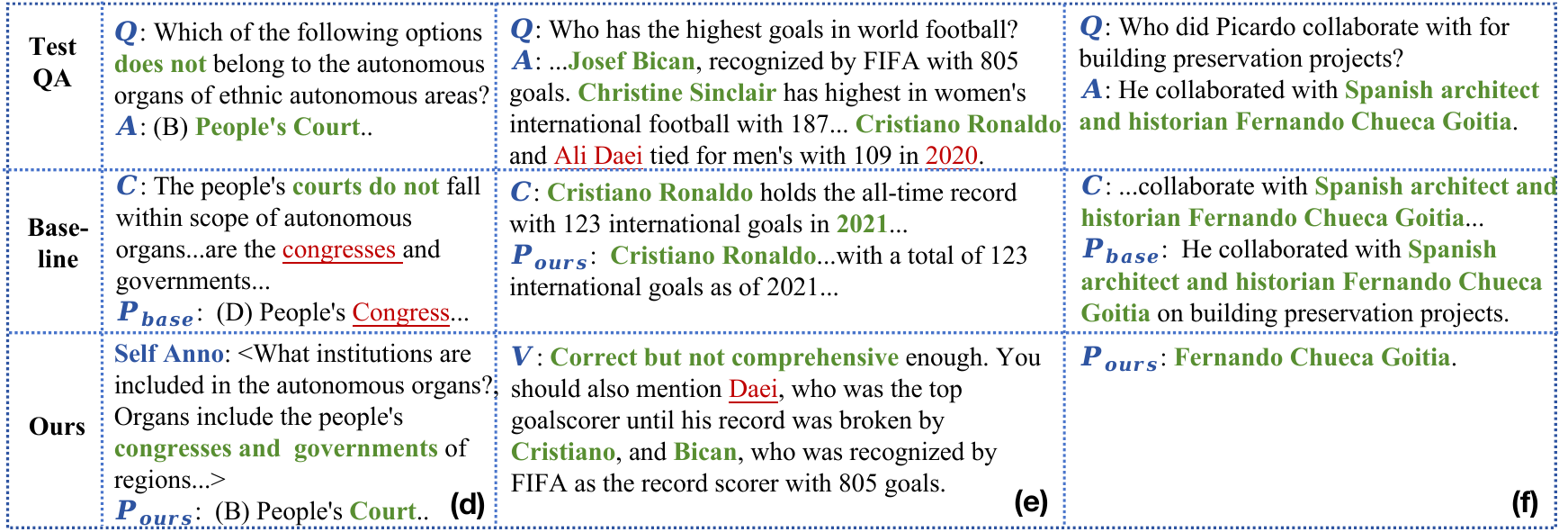}
\caption{More cases for KBAlign and baseline comparison. }
\label{fig:supcase}
\end{figure*}

We provide several more cases in Fig.~\ref{fig:supcase}. Case (d) shows that for different forms of tasks such as multiple choices, the self-annotated data can also provide key knowledge for the model. Case (e) shows a verification example in which the error can only be described in explicit natural language instead of a wrong label. Case (f) shows that our method does not always help improve the performance. In this case, the model discards some useful information due to the concise language style bias.

\subsection{General Domain Performance}

We conduct experiments on the widely adopted MMLU benchmark~\citep{hendryckstest2021} to evaluate the general ability reduction of KBAlign brought to the backbone model. \textit{w} mix refers to mixing general instruction tuning data~\citep{ding2023enhancing} with specific KB data. As shown in Table~\ref{tab:balance}, tuning on mixed data could also achieve most of the downstream improvement with a fairly small general performance degradation.

\begin{table}[ht]
\centering
\resizebox{0.7\linewidth}{!}{
\begin{tabular}{lcc}
\hline
\textbf{Methods}      & \textbf{MMLU} & \textbf{LooGLE F1} \\ 
\hline
\multicolumn{3}{c}{\textbf{MiniCPM-2B}} \\
\hline
Vanilla RAG & 44.07 & 30.92\\
\textbf{Ours} & 38.54 & 54.09\\
\ \ \ \textit{w} mix & 43.45 & 52.84\\
\hline
\multicolumn{3}{c}{\textbf{Llama3.1-8B-Instruct}}  \\
\hline
Vanilla RAG & 57.38 & 40.46\\
\textbf{Ours} & 49.41 & 62.07\\
\ \ \ \textit{w} mix & 54.61 & 61.33 \\
\hline

\hline
\end{tabular}}
\caption{Balance between domain knowledge and general capacity.}
\label{tab:balance}
\end{table}

\begin{algorithm}[t]
\caption{KBAlign Framework}
\label{alg:qa_train}
\begin{algorithmic}[1]
\Require Model $M$, Retriever $R$, Golden context $C_g$, Question $Q$, Answer $A$, Split size $k$

\Ensure Fine-tuned model $M_k$

\State \textbf{Annotation Process:}
\Procedure{ShortAnnotation}{\textnormal{$C_g$}}  % 修复参数传递问题
    \State $Q_{\mathrm{short}} \gets M(C_g)$
    \State $C_R \gets R(Q_{\mathrm{short}})$
    \State $C \gets C_g \oplus C_R$
    \State $A_{\mathrm{short}} \gets M(Q_{\mathrm{short}} \oplus C)$
    \State \Return $\langle Q_{\mathrm{short}}, A_{\mathrm{short}} \rangle$
\EndProcedure

\Procedure{LongAnnotation}{\textnormal{$\{S_i\}_{i=1}^n$}}  % 修复参数传递问题
    \State $C_g \gets \bigoplus_{i=1}^n S_i$
    \State $Q_{\mathrm{long}} \gets M(C_g)$
    \State $C_R \gets R(Q_{\mathrm{long}})$
    \For{$i = 1,\ldots,n$}
        \State $C_i \gets S_i \oplus C_R$
        \State $I_i \gets M(Q_{\mathrm{long}} \oplus C_i)$
    \EndFor
    \State $A_{\mathrm{long}} \gets M(Q_{\mathrm{long}} \oplus \bigoplus_{i=1}^n I_i)$
    \State \Return $\langle Q_{\mathrm{long}}, A_{\mathrm{long}} \rangle$
\EndProcedure

\State \textbf{Training Phase:}
\State Split annotated data $\{\langle Q, A \rangle\}$ into $k$ parts $\{\langle Q_i, A_i \rangle\}_{i=1}^k$

\State \textbf{Initial Tuning:}
\State $\mathcal{L}_1 = 0.5\mathbb{E}[\|M(Q_1) - A_1\|] + 0.5\mathbb{E}[\|M(Q_1 \oplus R(Q_1)) - A_1\|]$
\State $M_1 \gets \arg\min_{M} \mathcal{L}_1$

\State \textbf{Iterative Verifying:}
\For{$i = 2$ \textbf{to} $k$}
    \State $C_R \gets R(Q_i)$
    \State $P_i \gets M_{i-1}(Q_i \oplus C_R)$
    \State $V_i \gets M_{i-1}(Q_i \oplus P_i \oplus A_i)$
    
    \State $\mathcal{L}_i = 0.375\mathbb{E}[\|M(Q_i) - A_i\|] + 0.375\mathbb{E}[\|M(Q_i \oplus C_R) - A_i\|]$ 
    \State $\quad + 0.125\mathbb{E}[\|M(Q_i \oplus P_i) - V_i\|] + 0.125\mathbb{E}[\|M(Q_i \oplus C_R \oplus P_i) - V_i\|]$
    
    \State $M_i \gets \arg\min_{M_{i-1}} \mathcal{L}_i$
\EndFor
\end{algorithmic}
\end{algorithm}

\end{document}